\documentclass[sigconf]{acmart}

\renewcommand\footnotetextcopyrightpermission[1]{} 

\usepackage{booktabs} 
\usepackage[ruled,vlined]{algorithm2e}

\newcommand{\argmin}{\arg\!\min}

\acmConference[ADKDD'18]{ACM conference}{August 2018}{London, United Kingdom}
\begin{document}
\title{Dynamic Hierarchical Empirical Bayes: A Predictive Model Applied to Online Advertising}

\author{Yuan Yuan}
\affiliation{}
\email{yuyuan@adobe.com}

\author{Xiaojing Dong}
\affiliation{}
\email{xdong1@scu.edu}

\author{Chen Dong}
\affiliation{}
\email{chedong@adobe.com}

\author{Yiwen Sun}
\affiliation{}
\email{yisun@adobe.com}

\author{Zhenyu Yan}
\affiliation{}
\email{wyan@adobe.com}

\author{Abhishek Pani}
\affiliation{}
\email{apani@adobe.com}

\renewcommand{\shortauthors}{}
\renewcommand{\shorttitle}{Dynamic Hierarchical Empirical Bayes}

\begin{abstract}
Predicting keywords performance, such as number of impressions, click-through rate (CTR), conversion rate (CVR), revenue per click (RPC), and cost per click (CPC), is critical for sponsored search in the online advertising industry. An interesting phenomenon is that, despite the size of the overall data, the data are very sparse at the individual unit level. To overcome the sparsity and leverage hierarchical information across the data structure, we propose a Dynamic Hierarchical Empirical Bayesian (DHEB) model that dynamically determines the hierarchy through a data-driven process and provides shrinkage-based estimations. Our method is also equipped with an efficient empirical approach to derive inferences through the hierarchy. We evaluate the proposed method in both simulated and real-world datasets and compare to several competitive models. The results favor the proposed method among all comparisons in terms of both accuracy and efficiency. In the end, we design a two-phase system to serve prediction in real time.
\end{abstract}

%
%
%

\keywords{Hierarchical Bayes, Empirical Bayes, Hierarchy Determination, Hierarchical Shrinkage Loss, Dynamic Hierarchical Empirical Bayes, Sponsored Search, Revenue Per Click Prediction}

\maketitle

\section{Introduction}

An interesting phenomenon in many resource allocation decisions in marketing is that, at the decision unit level, the data are still very sparse, despite the size of the overall data. To leverage the rest of the information in the big data, hierarchical Bayes (HB) \cite{Gelman, Rossi} provides a natural solution by statistically borrowing information with a shrinkage-based estimation at the individual unit level. There are two challenges when applying a HB model. First, the hierarchy structure needs to be determined in advance, which could be a challenge especially when the data do not possess a clear hierarchical affiliation relationship. Second, in practice, researchers tend to use only two or three levels for HB models because, for a fully Bayesian analysis, simulation-based approaches are necessary to obtain the joint posterior distribution. If there are too many levels, the model could be computationally expensive and very sensitive to the distribution assumptions and priors when applied to real-world data in order to converge. 

In this paper, we develop a new model that dynamically determines the hierarchy based on the input data. Meanwhile, by adopting empirical Bayes \cite{Casella}, we present an empirical approach to get inferences through the hierarchical structure. We show a two-phase system where flexible multi-level hierarchical models with deep hierarchy can be applied efficiently. Inspired by the loss concept in tree models (e.g., CART \cite{Breiman}), we propose a Dynamic Hierarchical Empirical Bayesian (DHEB) method that is capable of dynamically constructing the hierarchy. Specifically, each sub region in a layer of the hierarchy is treated as a node. The challenge is to find a natural way to merge the idea of loss function into the HB framework so that the estimates derived by the HB model are consistent with the optimal solutions for the loss function. To do so, we propose a loss function with a regularization term that incorporates the Bayesian concept of prior \cite{Rasmussen}. More details can be found in section 4.2. Given the loss function, instead of a fully Bayesian analysis, we present a stepwise method that practices empirical Bayes and builds a hierarchy dynamically from top to bottom. This proposed methodology combines the advantages of both (1) the hierarchical Bayesian model, which allows information borrowing from similar branches, and (2) a tree model, which helps define the structure using data.

The performance of the proposed method is evaluated using a set of simulated data and real-world data from Adobe Advertising Cloud. We compare the proposed method with baseline models: weighted average, regularized linear regression, and fully HB models with different levels. All of the comparisons favor the proposed method against all its competitors in terms of prediction accuracy and efficiency.

The rest of the paper is organized as follows. We will first describe the process of the sponsored search and the challenges faced when evaluating ads performance in section 2, followed by some related work in section 3. We then introduce our proposed method in section 4. Section 5 and section 6 provide the simulation and experimental results. Finally, we conclude the paper in section 7.

\section{Background}
\subsection{Sponsored Search}
Sponsored search advertising is a kind of auction-based keyword advertising in search engines \cite{Lahaie}. Search engines decide the winners of the auctions based on their expected revenue. Meanwhile, advertisers need to understand what keywords are more valuable using performance measurements, such as number of impressions, click-through rate (CTR), conversion rate (CVR), revenue per click (RPC), cost per click (CPC), etc., so that they can manage their bids efficiently and allocate their budgets accordingly. Here, revenue is defined by advertisers' goals, which can be dollar revenue, number of orders, number of subscriptions, and so on. The winning ads are charged by user clicks, meaning that advertisers only pay when their ads are clicked by users.

Search engines provide platforms for advertisers to manage their bids and apply targeting and budgeting decisions. Figure 1 illustrates a typical hierarchical structure of bid management. Advertisers first create an account and construct several campaigns in the account. For each campaign, advertisers can group keywords and ads into ad groups for targeting and management purposes. Ads are often shared by keywords in a common ad group. For each keyword, advertisers can also determine the matching types used between keywords and search queries, such as ``broad match,'' ``exact match,'' and ``other'' match types. Advertisers can set targeting criteria using geographic and demographic information at the ad group or campaign level.
\begin{figure}
\includegraphics[height=1.5in, width=2.5in]{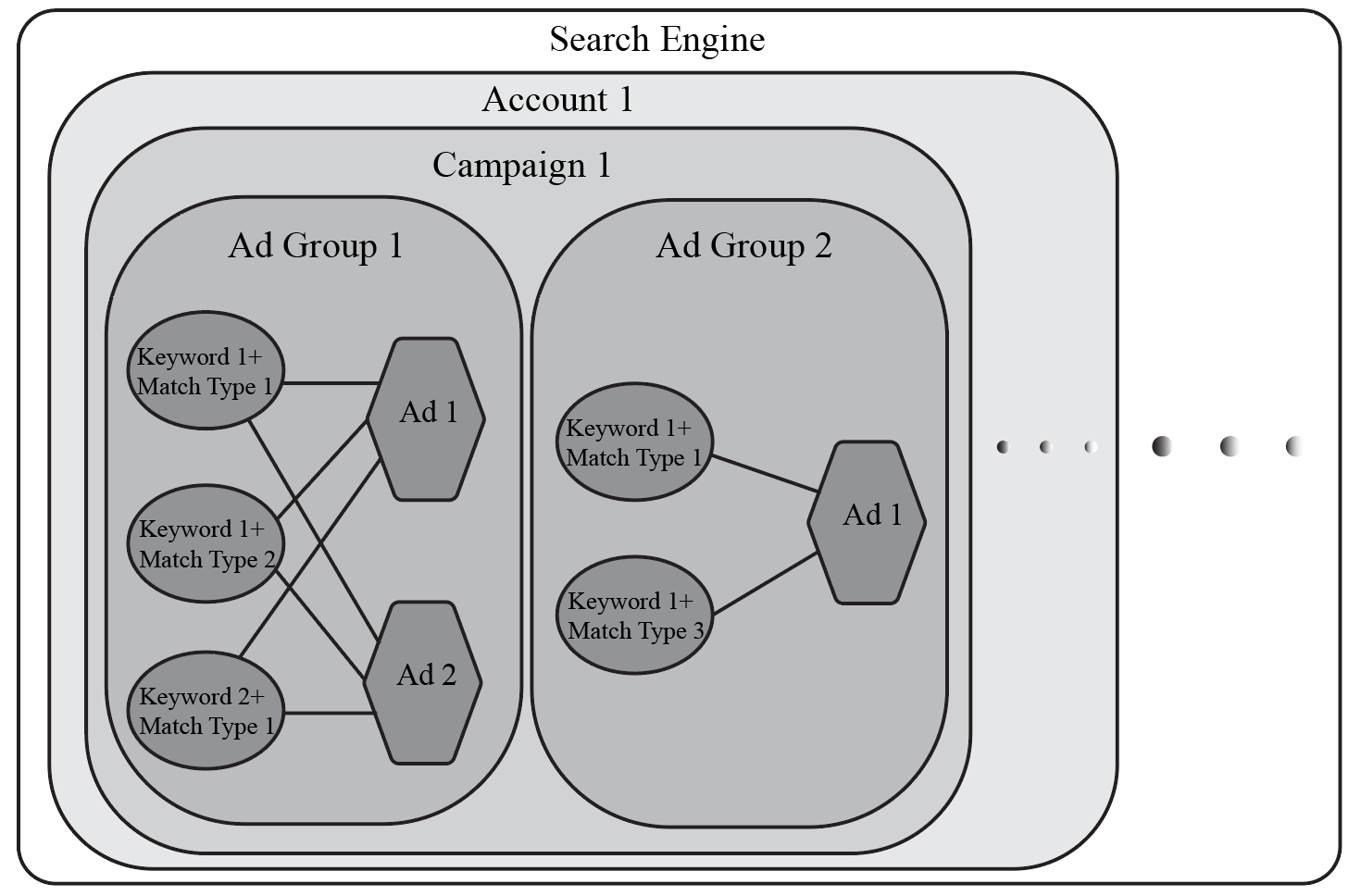}
\caption{Hierarchical structure of bid management.}
\end{figure}

\subsection{RPC Prediction}
In this paper, we focus on RPC prediction from the advertisers' perspective. First, we define ``bid units'' as the atomic units at which advertisers set their bids. Bid units are different from keywords because the same keywords can be targeted in multiple ad groups or campaigns and set with different bids. For example, in Figure 1, we consider ``Keyword 1 + Match Type 1'' under ``Ad Group 1'' as a bid unit and ``Keyword 1 + Match Type 1'' under ``Ad Group 2'' as another bid unit. The performance data we collect on the advertisers' side contain daily impressions, clicks, conversions and attributed revenue at the bid-unit level, and we remove the records with zero clicks because our goal is to predict the RPC for each bid unit. The problem is that, given the historical clicks and revenue data $\{x_{i,m},y_{i,m}; m=1,2,...,n_i\}$, we want to predict the next day's RPC for bid unit $i$. The features we can utilize are the hierarchical structure information of the bid units, such as corresponding campaigns, ad groups, and keywords, as well as some upper level variables. Here, upper level variables refer to the information above the bid-unit level, such as geo targeting at the campaign level, which is shared by the bid units under each campaign.

A well-known challenge in the RPC prediction problem is that, at the bid-unit level, the data are very sparse. From the perspective of users' behaviors, the sparsity challenge is twofold. First, for a large number of bid units, only a small number of days record non-zero clicks. We name the sparsity of clicks as $x$-sparsity. Second, among all the bid units that are clicked, the majority does not generate any revenue for the advertiser. This sparsity of revenue is denoted as $y$-sparsity. To further illustrate this phenomenon, we examine one month of data for a client of Adobe Advertising Cloud. The average $x$-sparsity and $y$-sparsity are about 90\% and 98\%, meaning only 10\% of the dates collect click data and among the dates with click data, about 98\% have zero revenue. Thus, if we build models at the bid-unit level by pushing down the upper level variables, we tend to generate zero RPC predictions for most bid units. These sparse predictions are undesirable for online advertising for several reasons. First, the bid units have potentials. Previous records of value zero do not necessarily mean the following day still bears a zero, and these potentials would be fully ignored by sparse predictions, leading to an overfitting model. Second, sparse predictions do not help distinguish the bid units if limited resources need to be allocated to them. 
\begin{figure}
\includegraphics[height=2in, width=3in]{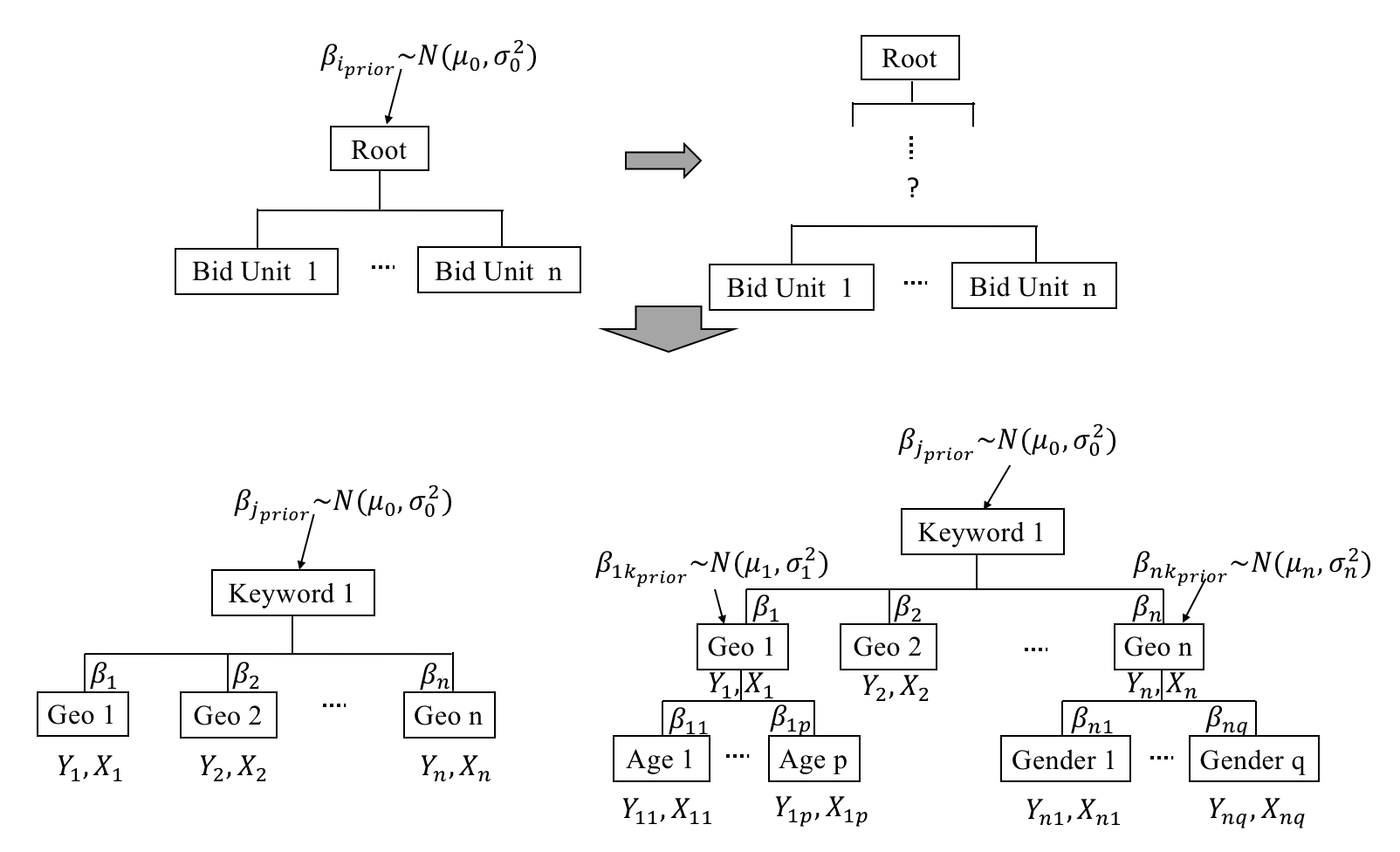}
\caption{Dynamic hierarchy choosing process.}
\end{figure}
\section{Related Work}
Although RPC is a vital metric in advertiser bidding decisions, the RPC-related literature is limited, partly because of the confidentiality of revenue data. Among the few existing studies, the work most related to our study is \cite{Sodomka}, which proposed a hierarchical model for predicting value per click, where the hierarchy is fixed a \textit{priori} and defined by ad group, campaign, and account. A linear model is used at each layer, and the aggregated loss is minimized. On the other hand, extensive literature has studied CTR and CVR predictions and  offered some attempts to utilize the data hierarchies/clusters in addressing data sparsity. Among those few attempts, \cite{Agarwal} assumed a predefined advertiser-publisher pair hierarchy and built a Poisson log-linear model for each node. Using the same data hierarchy, \cite{Agarwal2} proposed a tree-structured Markov model. Other than linear regression, \cite{Menon} modeled CTR from a collaborative filtering perspective of view. In addition to the preexisting advertiser hierarchy and publisher hierarchy, \cite{Lee} also considered clustering user-level information by grouping data within a specified Euclidean distance. 

To the best of our knowledge, all existing methods require a predetermined hierarchy, a \textit{priori} using the data structure and feature set in the data, which becomes a challenge when more user-defined features are involved. Our study provides a methodology that determines the hierarchical structure using information in the data so that the structure can be determined layer by layer during the model estimation process. Another contribution of our study to the literature is that the existing methods allow the child nodes to borrow information from their parents, mostly by combining the mean values of the parents and the children while ignoring the uncertainty of the mean values. In this paper, we propose a new method that allows the uncertainty to be incorporated before combining these values from parent or child nodes. 
\section{Methodology}
In this section, we present the proposed methodology in detail. For illustration, we first demonstrate how a two-level Bayesian regression model can be utilized in the RPC prediction problem in section 4.1. Then, we introduce hierarchical shrinkage loss (HSL) for determining the hierarchy empirically in section 4.2. We finish our discussion of the proposed DHEB method in section 4.3. 

\subsection{Two-level Hierarchical Bayes}
For each bid unit $i$, we denote its RPC $\beta_i$ as a random variable. Then we construct a linear regression model:
\begin{displaymath}
  Y_i=\beta_i X_i+\epsilon_i,
\end{displaymath}
where $X_i=(x_{i,1},x_{i,2}, ... ,x_{i,n_i} )^T$ and $Y_i=(y_{i,1},y_{i,2}, ... ,y_{i,n_i} )^T$ are historical number of clicks and revenue, respectively, and $\epsilon_i\sim N(0,\sigma_{\epsilon_i}^2)$. Our goal is to make an estimation for $\beta_i$ for each bid unit. Under the Bayesian framework, we assume a prior distribution of parameter $\beta_i$, then combine the prior with the likelihood function to yield a posterior. Assume $\beta_i$  has a normal prior distribution: 
\begin{displaymath}
  \beta_{i_{prior}}  \sim N(\mu_0,\sigma_0^2 ),
\end{displaymath}
where $\mu_0$, $\sigma_0$ are pre-specified hyper-parameters. Given the likelihood $Y_i |\beta_i,X_i,\sigma_{\epsilon_i}^2 \sim N(\beta_i X_i,\sigma_{\epsilon_i}^2 I_{n_i })$, where $I_{n_i}$ is an $n_i \times n_i$ identity matrix, the posterior for $\beta_i$ is:
\begin{displaymath}
  \beta_{i_{post}} | Y_i,X_i,\sigma_{\epsilon_i}^2\propto P(Y_i | \beta_i,X_i,\sigma_{\epsilon_i}^2 )P(\beta_{i_{prior}}) \sim N(\mu_i,\sigma_i^2 ),
\end{displaymath}
\begin{equation}
 \mu_i = \frac{(\sigma_0^2)^{-1} \mu_0+(\sigma_{\epsilon_i}^2 )^{-1} X_i^T X_i (X_i^T X_i )^{-1} X_i^T Y_i}{(\sigma_0^2)^{-1}+(\sigma_{\epsilon_i}^2)^{-1} X_i^T X_i},
\end{equation}
\begin{equation}
  (\sigma_i^2)^{-1}=(\sigma_0^2)^{-1}+(\sigma_{\epsilon_i}^2)^{-1} X_i^T X_i.
\end{equation}
By applying the same prior distribution for all $\beta_i$s and using the posterior mean as the predicted RPC for each bid unit, we get non-sparse predictions that contain information borrowed by incorporating a prior distribution. This prior information can be obtained by empirical Bayes leveraging the overall data. 

For data containing more features, a multi-level hierarchical Bayesian method is required to enable the propagation of information across the hierarchical structure and allow for information sharing among subgroups related in the hierarchy. For example, bid
units in the same ad groups may intuitively perform more similarly; thus, it makes more sense for them to share the same prior distribution. We fix the bottom level of the hierarchy to be the bid-unit level in order to differentiate the various bid units. The question now is determining the appropriate intermediate levels as shown in the top row of Figure 2. In a conventional hierarchical Bayesian model, the hierarchy is predetermined by domain knowledge. In our application, although there is a hierarchical structure for bid management as we introduced in section 2.1, issues still exist when trying to set the hierarchy involving features without a natural hierarchy. For example, under each ad group, advertisers set multiple keywords to target, which indicates that we can create a hierarchy with ``Keyword'' under ``Ad Group.'' Nevertheless, a common keyword can also appear in different ad groups targeting different types of customers. In this case, it is reasonable to put ``Ad Group'' under ``Keyword'' as well. This situation then calls for a data-driven approach to determine the hierarchy structure for the HB model.

\subsection{Hierarchical Shrinkage Loss}
Intuitively, it is similar to tree splitting using categorical variables in tree models, which grows a tree according to a certain predefined loss. In the interest of visualization and brevity, we use the terminology ``node'' as in tree models. The root node contains the whole population with all bid units. If we use ``Keyword'' as the first splitting feature and there are $n$ unique keywords in the data, then the root node will be split to $n$ child nodes, with each containing the bid units that share the same keyword. For bid units in each child node, we estimate the same RPC for them, and we assume that child nodes under a common parent node share the same prior distribution; thus, we use the term ``parent information'' to represent the ``prior information.'' Based on the observation of the posterior mean (1), which is a weighted average of parent information and information of itself, we develop the hierarchical shrinkage loss (HSL):
\begin{equation}
  L_p (l,\beta)=\sum_{k=1}^{n_l}h(\alpha_{p_{lk}} f(\beta_{p_{lk}},X_{p_{lk}},Y_{p_{lk}}) + \gamma_{p_{lk}} g(\beta_{p_{lk}},\beta_p)),
\end{equation}
where $p$ denotes the parent node; $p_l=\{p_{l1},p_{l2}, ... ,p_{ln_l}\}$ denotes the child nodes of $p$ when splitting by feature $l$; $\beta_{p_{lk}}$ and $\beta_p$ represent the RPC predictions in child node $p_{lk}$ and parent node $p$, respectively;  $X_{p_{lk}}$ and $Y_{p_{lk}}$ are the data in child node $p_{lk}$; $f$ and $g$ are functions measuring the within-node loss and loss to the parent node; $\alpha_{p_{lk}}$ and $\beta_{p_{lk}}$ represent the importance of the two losses; and $h(x)$ is a scalar function that transforms x to the order of interest. 

There are two terms in HSL: the first measures the weighted information loss within each child node, and the second considers the discrepancy between the estimators of the child node and the parent node. The estimator of each child node then considers not only the data within itself, but also the information of its parent, who also inherits from its parent according to the hierarchy. This additional hierarchy information leads to a more stable model as information from a larger subgroup is used.

\subsection{Dynamic Hierarchical Empirical Bayes}
In this section, we illustrate how DHEB builds a hierarchy using HSL. In the multi-level hierarchical Bayesian method, it is assumed that the parameters of the child nodes under the same parent node are from a common prior distribution and the prior information flows through the hierarchy. In a fully Bayesian analysis, a complete joint posterior distribution is generated according to the predetermined hierarchy, and simulations are usually applied to get inferences. This process can be computationally expensive. Instead of a fully Bayesian analysis, we employ empirical Bayes to grow the hierarchy from top to bottom. The proposed method not only provides a method for determining the hierarchy, but also presents an efficient way to get inferences.

We illustrate how to construct a loss function to choose the splitting features for the intermediate levels using an example in the bottom row of Figure 2. Suppose we are in the node ``Keyword 1'' and want to decide which feature to use for the further subdivision. Assume we use ``Geo'' as the splitting feature and split the data for each ``Geo'' as a child node $j$. Here, we use $j$ to differentiate from bid unit $i$ in section 4.1. Similar to section 4.1, we assume all RPCs $\beta_j$s under ``Keyword 1'' across different ``Geo's'' are related and generated from a common prior distribution, which is $\beta_{j_{prior}}  \sim  N(\mu_0,\sigma_0^2)$. Then the posterior distribution of $\beta_j$ for each ``Geo'' node is $\beta_{j_{post}} | Y_j,X_j,\sigma_{\epsilon_j}^2 \sim N(\mu_j,\sigma_j^2 )$, where
\begin{equation}
 \mu_j = \frac{(\sigma_0^2)^{-1} \mu_0+(\sigma_{\epsilon_j}^2 )^{-1} X_j^T X_j (X_j^T X_j )^{-1} X_j^T Y_j}{(\sigma_0^2)^{-1}+(\sigma_{\epsilon_j}^2)^{-1} X_j^T X_j},
\end{equation}
\begin{equation}
  (\sigma_j^2)^{-1}=(\sigma_0^2)^{-1}+(\sigma_{\epsilon_j}^2)^{-1} X_j^T X_j.
\end{equation}
Using the posterior mean $\mu_j$ as an estimate for $\beta_j$ in each child node, we can construct a loss function by degenerating (3) to the current layer as follows:
\begin{align}
 L_p (l ,\beta)=&\sum_{j=1}^n n_j ((\sigma_{\epsilon_j}^2 )^{-1} X_j^T X_j (\beta_j-(X_j^T X_j )^{-1} X_j^T Y_j )^2+\nonumber\\ 
&(\sigma_0^2)^{-1}  (\beta_j - \mu_0 )^2 ) \nonumber\\
=& \sum_{j=1}^n n_j (\alpha_j f(\beta_j,X_j,Y_j )+\gamma_j g(\beta_j,\mu_0 )),\label{eqn:1}
\end{align}
where the generic functions in (3) are \\
$f(\beta_j,X_j,Y_j )=(\beta_j-(X_j^T X_j )^{-1} X_j^T Y_j )^2$, $g(\beta_j,\mu_0 )=(\beta_j - \mu_0 )^2$, $\alpha_j=(\sigma_{\epsilon_j}^2 )^{-1} X_j^T X_j$, $\gamma_j=(\sigma_0^2)^{-1}$, and $h(x)=nx$, with node $p_{lk}$ denoted as $j$ for short.

The optimal solution of $\beta_j$ would be $\mu_j$. Function $f$ represents the difference between the parameters of the child nodes and the OLS estimations based on the sample data. Function $g$ measures the difference between the parameters of the child nodes and parent node, which is represented by the prior mean. The weights of the two losses $\alpha_j$ and $\gamma_j$ are inversely proportional to the variance of the OLS estimator and prior variance. The basic idea is intuitive: If the prior variance is larger, it provides noisier information regarding the $\beta_j$ estimates and, hence, its contribution is smaller than the case when the prior variance is smaller. Similarly, if the sample data are divergent and noisy, they will get less weight. $h(x)=nx$, where $n$ is the number of observations in the node. We multiply the loss for each child node by the number of observations in the node because we shrink the loss to one node level by $f$ and $g$. In order to make the losses for different splitting features comparable, we calculate the loss at the individual observation level and treat the loss at one node level as a representation for all the observations in this node.

Once we have the loss function, we can decide which feature to use for partition as 
\begin{equation}
  l^*=\argmin_l L_p (l,\hat{\beta})
\end{equation}

Suppose we choose ``Geo'' for the second level and we need to decide the splitting variable for the third level. We assume the posterior distribution of $\beta_j$ as the prior distribution of $\beta_{jk}$ under ``Geo $j$'' and apply the same method recursively, which is $\beta_{{jk}_{prior}}  \sim \beta_{j_{post}}$ (Figure 2, bottom row right). 

To get loss (6), both prior distribution of $\beta_j$ and regression variance $\sigma_{\epsilon_j}^2$ are assumed known; therefore, sample data should be used to get estimations. For prior distribution, only the parameters in the root node are necessary because the posterior of the parent node would be used as prior for its child nodes. Empirical Bayes can be applied when there is a lack of prior knowledge. Here, we give an example by using the sample mean as the prior mean and weighted sample variance as the prior variance: $\mu_0=\frac{\sum_m y_m}{\sum_m x_m}$, $\sigma_0^2=\frac{\sum_m x_m (\frac{y_m}{x_m} -\mu_0 )^2}{\sum_m x_m - 1}$, where $m$ denotes total historical data for all bid units. The variance $\sigma_{\epsilon_j}^2$ needs to be estimated in each node which can be given by: $\hat{\beta}_{j,OLS}=(X_j^T X_j )^{-1} X_j^T Y_j$, $\sigma_{\epsilon_j}^2=\frac{1}{n_j-1} (Y_j-\hat{\beta}_{j,OLS} X_j )^T (Y_j-\hat{\beta}_{j,OLS} X_j )$,  where $\hat{\beta}_{j,OLS}$ is the OLS estimator and $n_j$ is the number of observations for node $j$.

Another problem is when to stop splitting. Here, we propose a stopping criterion:
\begin{equation}
\frac{SSE(p_{l^*} )}{SSE(p)}>r,
\end{equation}
where $SSE(p)=||Y_p-\hat{\beta}_p X_p ||^2$ and $SSE(p_{l^*} )=\sum_{j\in p_{l^*}}||Y_j- \hat{\beta}_j X_j ||^2$, denoting the sum of squared errors for the parent node $p$ and child nodes $p_{l^*}$. This means a node will stop growing when the total sum of squared errors does not decrease by a certain ratio $1-r$. 

The final step would be attaching the bid-unit level to the bottom of the chosen hierarchy. The procedure loops the leaf nodes of the hierarchy and subdivides them into child nodes, with each node containing the data for a specific bid unit. 

The proposed DHEB also provides an approach to get inferences through a hierarchy. If we have a fixed hierarchy, we can apply equations (4) and (5) to get stepwise posterior distributions from the root to bottom levels and then obtain inferences.
\begin{algorithm}
\caption{Dynamic Hierarchical Empirical Bayes}
\label{DHEB}
Initialize a set of nodes to split $Q=\{root\}$\

       \For{node $p$ in $Q$} {
          get splitting variable $l^*$ and create child nodes $C = p_{l^*}$ according to (7)\\
          \If{stopping criterion (8) is satisfied}{
             $C = \varnothing$
          }
          \Else{attach $C$ to $p$}

          $Q = Q\backslash\{p\}\bigcup C$\
       }

    \For{leaf node $ln$ in leaves} {
       \For{bid unit $bu$ in $ln$} {
          create child node $c$ for $bu$ and attach it to $ln$\
       }
   }
\end{algorithm}

\section{Simulation Results}
\begin{figure}
\includegraphics[height=1.8in, width=3.4in]{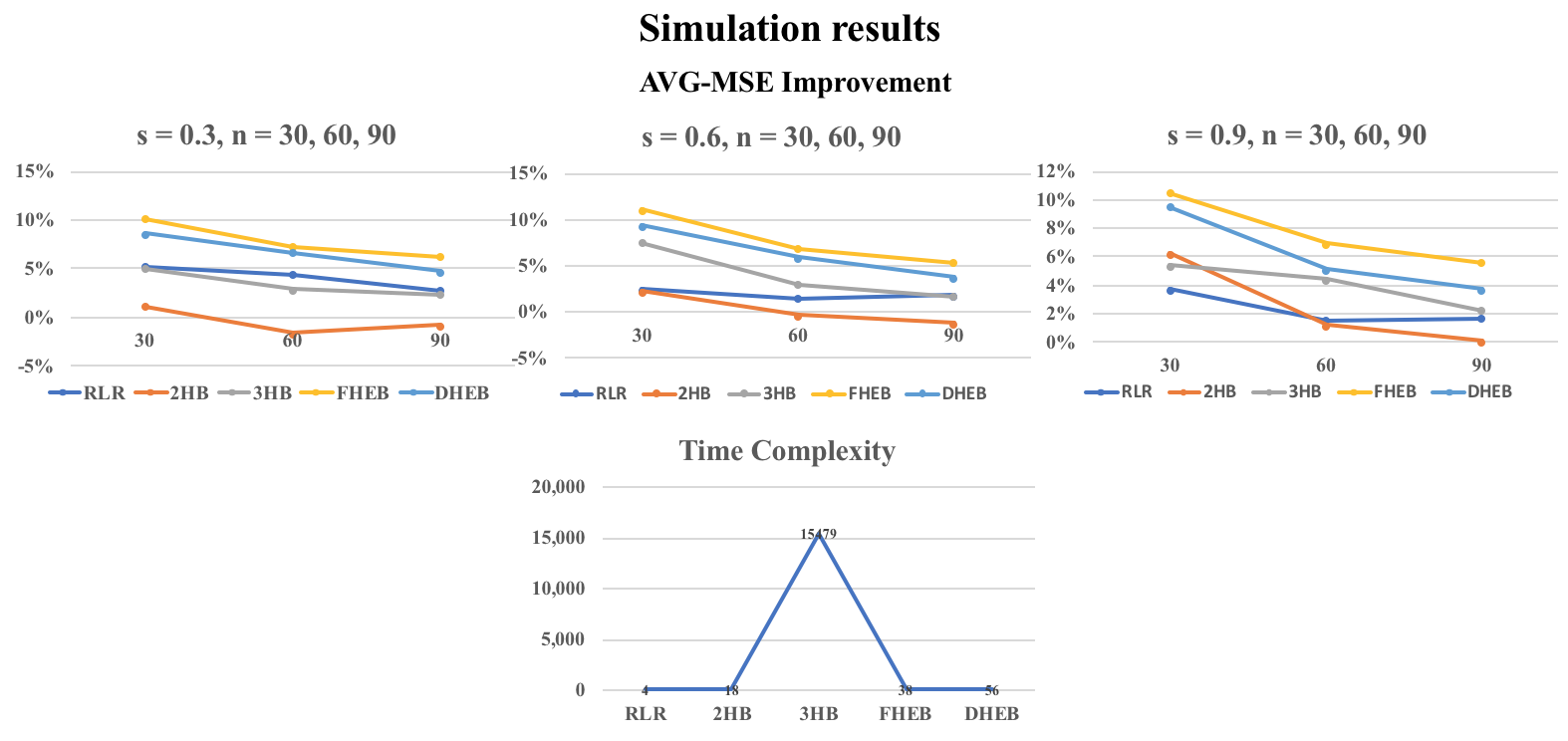}
\caption{Simulation results.}
\end{figure}
In this section, we evaluate the proposed method on several simulated datasets. On each dataset, we conduct an analysis using 6 models: 
\begin{enumerate}
\item Weighted average (WA): The predicted RPC of each bid unit is the weighted average of historical RPC using the number of clicks as weights.
\item Regularized linear regression (RLR): This fits a regularized linear regression by pushing down all the upper level features.
\item Two-level HB (2HB): This model was discussed in section 4.2. The hierarchy is ``Root - Bid Unit.''
\item Three-level HB (3HB): This model first predefines a three-level fixed hierarchy. Then we use Rstan \cite{Stan} to do the posterior sampling and treat the posterior means at the bottom level as predictions. We limit the hierarchy to three levels because the more levels we have, the more computationally expensive the model is and, when using real-world data, many assumptions may not be satisfied, which makes the simulation difficult to converge.
\item Multi-level fixed hierarchical empirical Bayes with true hierarchy (FHEB): We fix the hierarchy as the true hierarchy used during the simulation and apply the same inference approach as the proposed method.
\item DHEB: This is our proposed method.
\end{enumerate}

We apply these six models on a set of simulated data. Data are generated by the procedure as follows:
\begin{enumerate}
\item Create 100 bid units and 4 upper level features (named A, B, C, and D); each feature has 10 to 20 categories. Set the date range to 6 months (i.e., from ``2017-01-01'' to ``2017-06-30'').
\item Assume the implicit hierarchy is A - B - C - D - Bid Unit.
\item Set the top prior mean $\mu_0$ and variance $\sigma_0^2$.
\item For nodes in the intermediate  levels, generate the mean $\mu_c$ for child nodes from $N(\mu_p,\sigma_p^2 )$, which is the parent node distribution. The variance of child nodes $\sigma_c^2$ is predetermined. Here, we just use the same variance as $\sigma_0^2$.
\item For the bottom bid-unit level, we apply (4) to generate $\mu_c$ and set RPC for this bid unit as $\mu_c$. We generate a list of clicks $X$ with length $n$, then revenue $Y=RPC\times X+\epsilon$, where $\epsilon \sim N(0,\sigma_{\epsilon}^2)$ and $\sigma_{\epsilon}^2$ is predetermined.
\item $x$-sparsity is determined by $n$, which is the number of observations we generated in (5). The $x$-sparsity is higher as $n$ is smaller because we fix the date range. $y$-sparsity is denoted by $s$, and we will randomly set $s$ of the revenue to be zero. 
\item We apply 9 pairs of $n$ and $s$ combinations and generate 10 datasets for each combination.
\end{enumerate}

We use two months of data to predict RPCs for next day and test in a rolling way for 30 days. The performance metric is:
\begin{equation}
\text{AVG-MSE} =\frac{1}{N}(\sum_{t=1}^N \frac{1}{N_t} \sum_{i=1}^{N_t}(\hat{\beta}_{it} \times x_{it} - y_{it})^2),
\end{equation}
where $N=30$ is the number of testing days; $N_t$ is the number of bid units on day $t$; $(x_{it}, y_{it})$ is the true data for bid unit $i$ on day $t$; and $\hat{\beta}_{it}$ is the predicted RPC. For 3HB, we apply a hierarchy as ``Root - A - Bid Unit''; for FHEB, we deploy the true hierarchy. Because WA is a baseline method, we calculate the improvement of the other 5 models relative to WA. The improvement is represented by the reduced percentage of AVG-MSE and would be negative if the AVG-MSE of WA is smaller. The top row of Figure 3 shows the comparison of the 5 models for different $n$ and $s$. As we can see, for different combinations of $n$ and $s$, FHEB outperforms all the other models, with DHEB ranking second. As both $x$-sparsity and $y$-sparsity increase, the benefits obtained from FHEB and DHEB become greater. For time complexity, we plot the ratio between the running time of the other 5 models and WA, as shown in the bottom row of Figure 3. For 3HB, it takes a much longer time to do the sampling, and the model does not perform best due to the limited number of levels. In practice, if we are confident about what the true hierarchy is, FHEB provides an approach to get predictions without worrying about time complexity. If we are not sure about how to build the hierarchy, DHEB can determine the hierarchy empirically and give desirable predictions.
\section{Experimental Results}
\subsection{Model Comparison}
The data come from multiple online advertising campaigns owned by a common advertiser. In the bid-unit level, the performance data contain the number of clicks and collected revenue, ranging from ``2017-09-01'' to ``2017-12-23''. In addition, the data record the structural features of the keywords as shown in Table 1. Here, instead of regular hierarchical features, we introduce ``Day of Week,'' which provides an additional group for the daily data by indicating whether the day is Monday, Tuesday, etc. ``Geo'' represents geo targeting for the campaigns; it only has one unique category in this dataset. When a hierarchy is established in a hierarchical model, some features have a natural relationship with each other, such as ``Search Engine,'' ``Account,'' ``Campaign,'' and ``Ad Group.'' However, for ``Geo,'' ``Keyword,'' ``Match Type,'' and ``Day of Week,'' it is hard to determine their positions and order. In addition, it may not be necessary to include all structural features in the hierarchy.
\begin{table}
  \caption{Structural features statistics}
  \label{tab:commands}
  \begin{tabular}{ccl}
    \toprule
    Feature & Symbol & \# Categories\\
    \midrule
    Search Engine & SE & 3 \\
    Account & AC & 3\\
    Geo & GEO & 1\\
    Campaign & CP & 156\\
    Ad Group & AG & 2847\\
    Keyword & KW & 3208\\
    Match Type & MT & 3\\
    Day of Week & DOW & 7\\
    Bid Unit & BU & 5148\\
    \bottomrule
  \end{tabular}
\end{table}
We compare the 6 models and use the same evaluation metric as in section 5. For 3HB, we apply a hierarchy as ``Root - Campaign - Bid Unit''; for FHEB, we chose a hierarchy by domain knowledge, which is ``Root - Search Engine - Account - Campaign - Ad group - Keyword - Match Type - Day of Week - Bid Unit.'' We dropped ``Geo'' because it is unique. In Figure 4, the left plot shows the AVG-MSE improvement compared with WA. As we can see, DHEB outperforms other methods. The middle plot demonstrates the time complexity compared with WA. 3HB takes a much longer time due to the sampling process. 
\begin{figure}
\includegraphics[height=0.9in, width=3.2in]{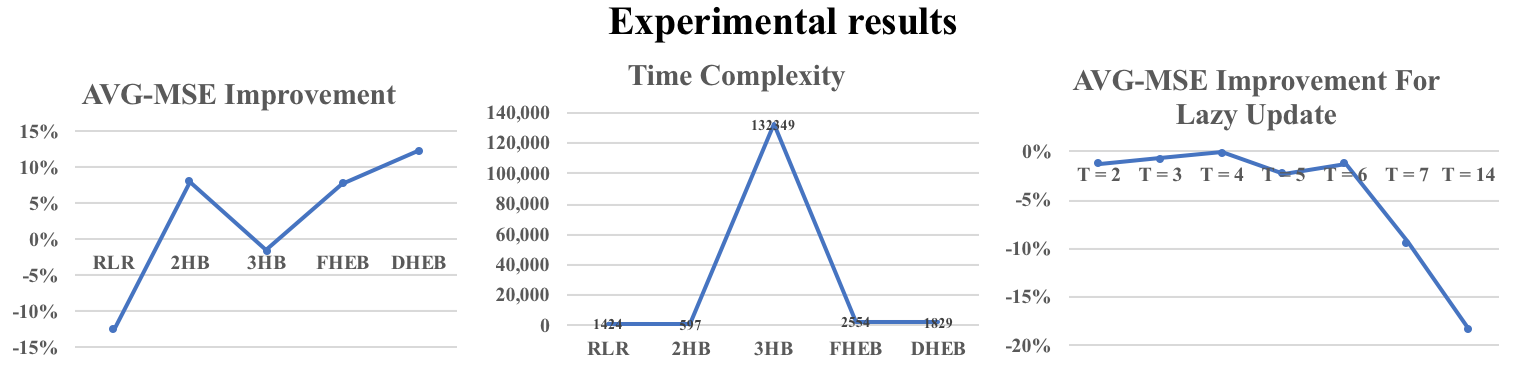}
\caption{Experimental results.}
\end{figure}
\begin{figure}
\includegraphics[height=1.5in, width=3.2in]{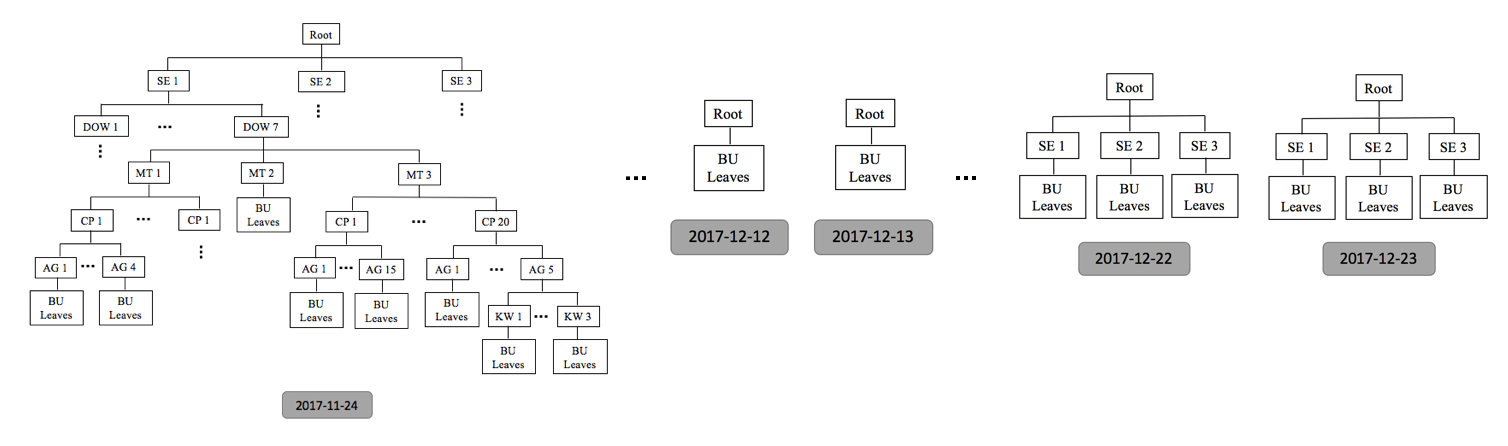}
\caption{Chosen hierarchy.}
\end{figure}
\subsection{Two-phase System}
\begin{figure}
\includegraphics[height=1.5in, width=2in]{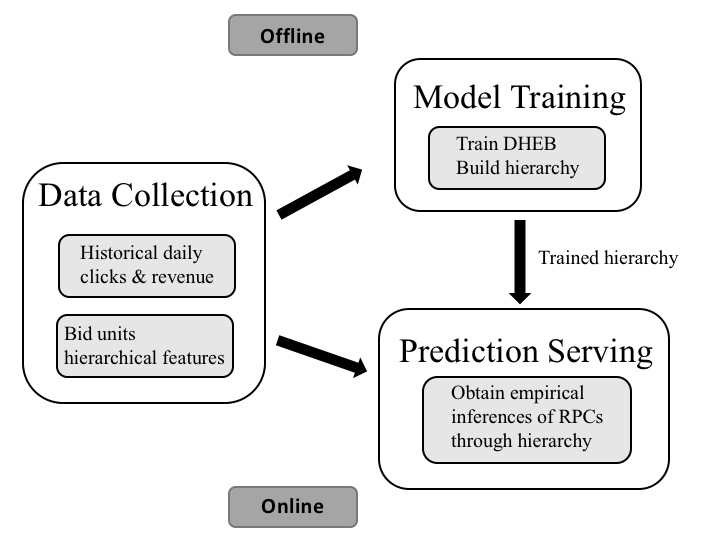}
\caption{Two-phase system.}
\end{figure}
Figure 5 shows the hierarchy trained on several testing days. As we can see, the hierarchy does not change frequently over time, which makes sense as there is only one day difference between the training data for two consecutive days. There are three modules: (1) data collection: obtaining bid units features, the daily number of clicks, and revenue in history; (2) model training: training the DHEB model and building a hierarchy; (3) prediction serving: giving RPC prediction based on the hierarchy determined in model training. Module (2) is the most time-comsuming part. We separate these three modules into offline and online phases as shown in Figure 6, where in offline phase, we do model training and in online phase, we do prediction serving based on the trained hierarchy. Given the fact that the hierarchy determined by DHEB does not change a lot in a short period, we schedule the offline phase in a low frequency and run the online phase in real time. We introduce parameter $T$ as the period of the offline phase. $T=1$ means we run the offline phase every day, $T=2$ means every other day, and so on.

The right plot in Figure 4 consists of AVG-MSE improvement compared to $T=1$ for different values of $T$. An appropriate $T$ would be 4, which will reduce the time complexity without making many sacrifices in model accuracy.
\section{Conclusions}
In this paper, we propose a Dynamic Hierarchical Empirical Bayesian (DHEB) method to build a multi-level hierarchical model to overcome the sparsity challenge in online advertising data. The proposed method provides a way to choose hierarchical levels by incorporating a loss function, such as the function used in tree models. The method is also equipped with an empirical Bayesian approach to get inferences through a hierarchy. It is applicable in many practical problems where data are sparse and hierarchical structure can be leveraged to obtain shrinkage-based estimations. In addition, the proposed regularized loss function can be applied in traditional tree models as well as other tree-based methods, as an approach to borrow information from the parent node in order to deal with data sparseness. We also present a two-phase system which can serve prediction in real time.

\bibliographystyle{ACM-Reference-Format}
\bibliography{sample-bibliography}

\end{document}